\documentclass{article}

\usepackage[utf8]{inputenc} 
\usepackage[T1]{fontenc}    
\usepackage{hyperref}       
\usepackage{url}            
\usepackage{booktabs}       
\usepackage{amsfonts}       
\usepackage{nicefrac}       
\usepackage{microtype}      
\usepackage{xcolor}         
\usepackage{amsmath}
\usepackage{cleveref}

\usepackage[round]{natbib}

\usepackage[final]{neurips_2022}

\usepackage{wrapfig,lipsum,booktabs}

\usepackage{lipsum}
\usepackage{mwe}

\usepackage[skip=0pt]{caption}
\usepackage[skip=0pt]{subcaption}
\usepackage{graphicx}

\usepackage[font=small]{caption}  

\usepackage{todonotes}

\title{Improving Dense Contrastive Learning with\newline Dense Negative Pairs}

\author{
  Berk Iskender$^{\dagger}$\thanks{Correspondence to berki2@illinois.edu and maryamkhademi@google.com}\\
    UIUC\\
\And Zhenlin Xu\thanks{Work done during an internship at Google}\\
Amazon\\
\And Simon Kornblith\\
 Google Research\\
\And En-Hung Chu\\
 Google\\
\And Maryam Khademi\\
  Google\\
  }

\begin{document}

\maketitle

\begin{abstract}
Many contrastive representation learning methods learn a single global representation of an entire image. However, dense contrastive representation learning methods such as DenseCL~\citep{wang2021dense} can learn better representations for tasks requiring stronger spatial localization of features, such as multi-label classification, detection, and segmentation. In this work, we study how to improve the quality of the representations learned by DenseCL by modifying the training scheme and objective function, and propose DenseCL++.
We also conduct several ablation studies to better understand the effects of: (i) various techniques to form dense negative pairs among augmentations of different images, (ii) cross-view dense negative and positive pairs, and (iii) an auxiliary reconstruction task. Our results show 3.5\% and 4\% mAP improvement over SimCLR \citep{chen2020simple} and DenseCL in COCO multi-label classification. In COCO and VOC segmentation tasks, we achieve 1.8\% and 0.7\% mIoU improvements over SimCLR, respectively. 

\end{abstract}

\section{Introduction}
\label{sec:intro}

Self-supervised learning aims to learn representations from unlabeled data via pre-text task training. Contrastive learning, as a self-supervised learning technique, performs the pre-text task of instance discrimination \citep{dosovitskiy2014discriminative, wu2018unsupervised, chen2020simple}. Instance discrimination in contrastive learning usually trains a single global representation, and these representations are principally evaluated in terms of downstream performance on a  single-label classification task. However, methods that perform well in this setting may perform suboptimally on multi-label classification tasks, where each label is associated with a distinct object in an image, but different image regions contain different semantic content. Motivated by the important application of multi-label classification in industry, we target representation learning for this task. We  demonstrate that our approach also improves accuracy on dense downstream tasks such as segmentation.

Our work is inspired by DenseCL \citep{wang2021dense} which proposes to use dense features rather than global ones in contrastive learning to improve the performance in dense prediction tasks. We focus on further boosting the performance of dense contrastive learning by modifying the training scheme and the objective function. Unlike DenseCL, our proposed approach formulates negative pairs between the dense features of augmented views of different images and uses their similarities in the proposed dense contrastive loss scheme. We show that the proposed method outperforms DenseCL in various settings. We also conduct several ablation studies to better understand the effects of: (i) various methods to form dense negative pairs among augmentations of different images, (ii) cross-view dense negative and positive pairs, and (iii) an auxiliary reconstruction task.
\section{Related Work}
\label{sec:related_work}

SimCLR \citep{chen2020simple} proposes a simple contrastive learning framework such that the \textit{projected} representations of randomly augmented views of the same image sample are attracted to each other using a contrastive loss. DenseCL \citep{wang2021dense} proposes an extension of this 
framework better suited to dense prediction tasks. In DenseCL, the contrastive loss is applied in a dense pairwise manner which improves the performance compared to the global representation learning counterparts \citep{chen2020improved}.

On the other hand, following widespread success of the transformer architecture \citep{vaswani2017attention} in NLP tasks, Vision Transformer (ViT) \citep{dosovitskiy2020image} adapts the architecture for visual tasks and achieves impressive results when pretrained on sufficient amount of data. Inspired by \citep{devlin2018bert}, \citet{li2021mst} explored the idea of introducing a reconstruction task to the contrastive learning framework using ViT as an encoder. 
\citet{wang2022repre} further study the use of reconstruction as a pretext task, incorporating a decoder module in various self-supervised contrastive settings. These two methods use shallow convolutional networks for their decoders to preferably learn additional useful local features in the latent space. However, the authors suggest that the use of sophisticated reconstruction models may be harmful to transfer tasks, as they could lead to excessively local representations.

\section{Method}
\label{sec:method}

\begin{figure*}[t]
    \centering
    \includegraphics[scale=0.5,trim={0 18pt 0 0},clip]{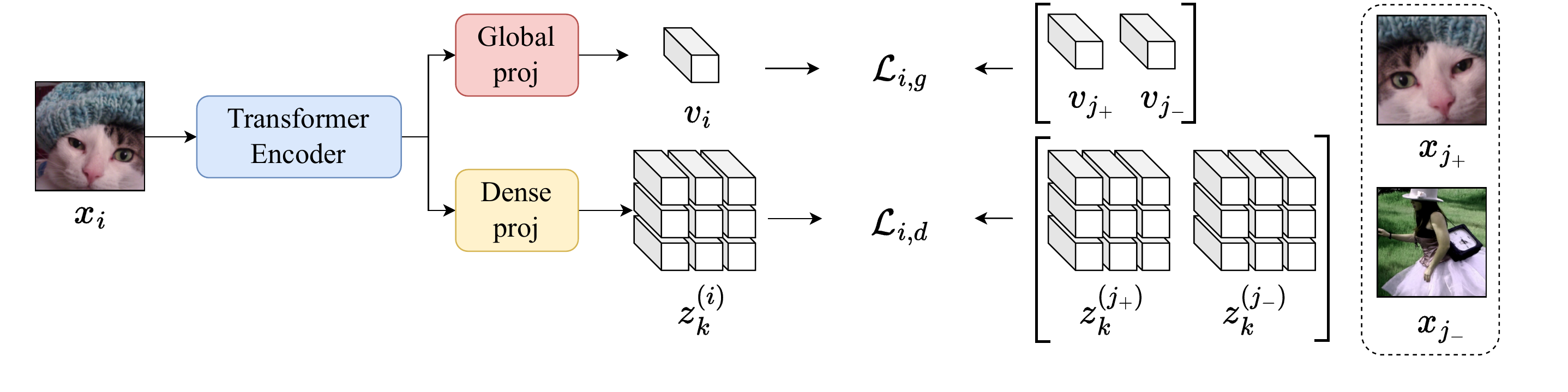}
    \caption{\small DenseCL++ training scheme. Global and dense positive/negative correspondences are used in the global (top row) and dense (bottom row) loss functions, respectively.}
    \vspace{0.25cm}
    \label{fig:denseclv1_diagram}
\end{figure*}

\subsection{Dense Contrastive Learning}
\label{sec:DenseCL}
Contrastive learning learns latent representations of signals for which the positive correspondences are attracted to each other and negative ones are repelled from one another.
Dense contrastive learning \citep{wang2021dense} further adapts this framework 
for dense prediction tasks by replacing the global representations with their dense counterparts. For each image, instead of the global representation $v \in \mathbb{R}^D$, $S \times S$ many dense feature vectors  $z \in \mathbb{R}^L$ are extracted.

Then, dense positive pairs are formed between dense features of the anchor view $x_i$ and its corresponding augmented view $x_{j_+}$ by finding the most similar correspondence for each dense vector of $x_i$ in $x_{j_+}$ as $k_+ = \arg\max_{l} \mathrm{sim}(z^{(i)}_{k}, z^{(j_+)}_{l})$,
where $z_k^{(i)}$ is the $k$-th dense feature of the anchor view $x_i$, $z_l^{(j_+)}$ is the $l$-th dense feature of $x_{j_+}$, and $\mathrm{sim}(a,b)$ calculates the cosine similarity between two feature vectors. Dense negative pairs are formed between the dense feature vectors of the anchor view and the global representations of views from other images. The dense contrastive loss is computed as
\begin{equation}
    \small \mathcal{L}_{i, d} = \sum_{k} -\log \frac{ \exp{(z_k^{(i)} \cdot z_{k_+}^{(j_+)} ) }/\tau }{ \exp{(z_k^{(i)} \cdot z_{k_+}^{(j_+)})} + \sum_{ j_- } \exp{ (z_k^{(i)} \cdot v_{j_-} ) / \tau } }
\end{equation}
where $z_{k_+}^{(j_+)}$ is the positive dense correspondence for the dense feature vector $z_k^{(i)}$ in the view $x_{j_+}$, $v_{j_-}$ is the global feature for the image $x_{j_-}$, and $\tau$ is the temperature parameter.

The overall loss is a linear combination of the global InfoNCE loss term $\mathcal{L}_{i,g}$ \citep{oord2018representation} and dense loss, $\mathcal{L}_i = (1-\lambda) \mathcal{L}_{i, g} + \lambda \mathcal{L}_{i, d}$,
where $\lambda \in [0,1]$ is a weight constant.

\subsection{DenseCL++: DenseCL with Dense Negative Pairs}
In our proposed method, instead of computing dense-global negative correspondences in dense contrastive loss $\mathcal{L}_{i, d}$, we form dense negative pairs. This leads to the following dense contrastive loss
\begin{equation}
\label{eq:densecl_w_dense_neg_pairs}
    \small \mathcal{L}_{i, d} = \sum_{k} -\log \frac{ \exp{(z_k^{(i)} \cdot z_{k_+}^{(j_+)} ) }/\tau }{ \exp{(z_k^{(i)} \cdot z_{k_+}^{(j_+)})} + \sum_{j_-, m} \exp{(z_k^{(i)} \cdot z_{m}^{(j_-)}) / \tau} },
\end{equation}
where the global representation $v_{j_-}$ of the view $x_{j_-}$ is replaced by its dense features $z_m^{(j_-)}$. This is illustrated in Fig. \ref{fig:denseclv1_diagram}.

\begin{figure}[t]
    \centering
    \begin{subfigure}{.45\textwidth}
    \centering
    \includegraphics[width=\linewidth,trim={0 25pt 0 20pt},clip]{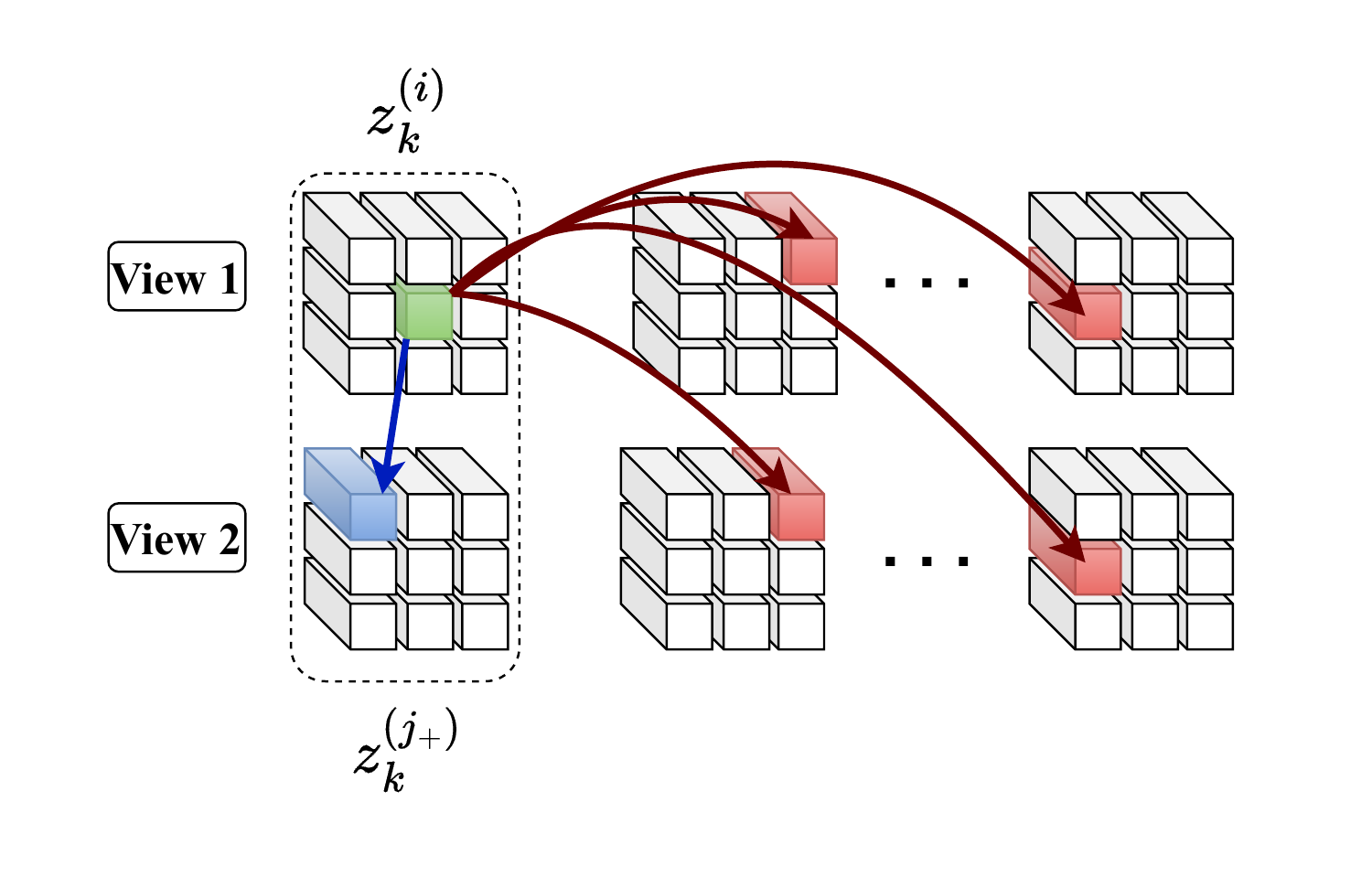}
    \caption{}
    \end{subfigure} 
    \begin{subfigure}{.45\textwidth}
    \centering
    \includegraphics[width=\linewidth,trim={0 25pt 0 20pt},clip]{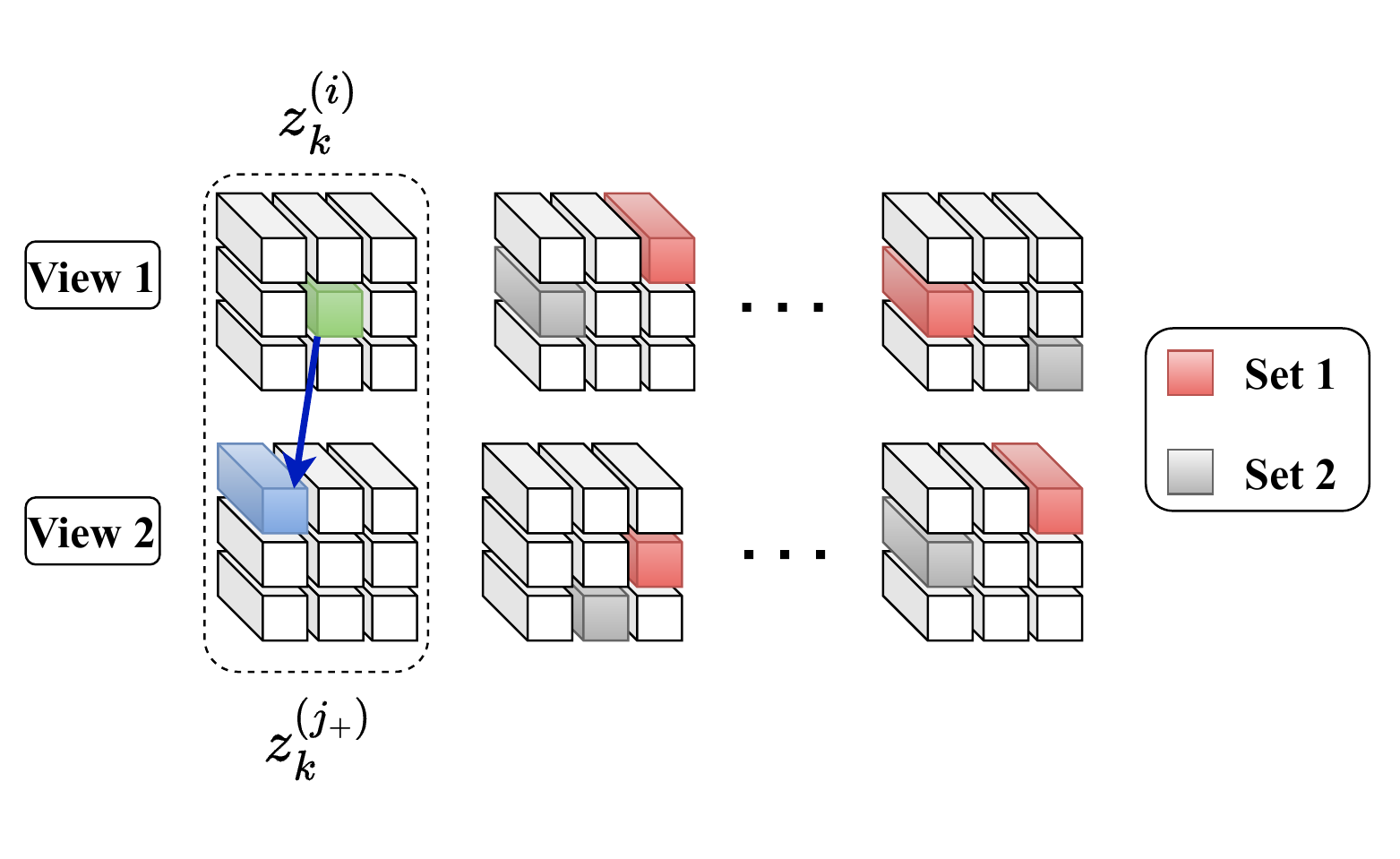}
    \caption{}
    \end{subfigure} 
    \caption{\small Dense positive and negative pair formulations for DenseCL++. Pair forming diagrams for (a) randomly sampling dense negatives, and (b) using $M=2$ candidate negative feature sets for guided sampling from other image augmentations.}
    \label{fig:densecl_dense_neg_diagrams}
\end{figure}

\label{sec:form_dense_neg_pairs}
\textbf{Forming dense negative pairs} Multiple possibilities exist for forming dense negative pairs. The first 
approach is to randomly sample a single dense feature from each augmented view in the batch and use them for the computation of negative pairs with the dense features of the anchor view. We use this option as the baseline of our proposed method. A diagram is provided in Fig. \ref{fig:densecl_dense_neg_diagrams} (a).

Another possibility is to sample $M$ random candidate sets of dense negative features where each set contains a single 
feature from each view in the batch of other images. Then, the similarities between each set and the dense features of the anchor image $x_i$ can be sorted using a specific criterion. Finally, the set with the top rank can be used to form dense negative pairs. The criterion can be chosen as the average similarity of the dense features in the set to all dense anchor features, with the hope that such sets include harder negatives.
We call this alternative the guided dense negative formulation.

To more easily distinguish sets with hard dense negatives, thresholding can be applied to the similarities $q = \mathrm{sim}(a, b)$ before computing the average 
\begin{equation}
    \bar{q}=\begin{cases} -1, & q \leq \beta \\ q, & \text{otherwise},\end{cases}
\end{equation}
where $\beta \in [-1,1]$ is the threshold constant. 

Since the corresponding views of an image are only used for forming dense positive pairs, it can be possible to observe high mean cosine similarities between dense features of these views. 
To improve discriminability and obtain a wider 
distribution of cross view dense features, $N$ additional 
pairs can be formed from the corresponding augmented view $x_{j_+}$ by selecting the least similar features.
\section{Experiments}
\label{sec:experiments}

\label{sec:framework}
\textbf{Framework} We use ViT-S/16 \citep{dosovitskiy2020image} as the backbone encoder in our experiments. Since we aim to compare the pretraining schemes and objectives, we do not explicitly try to optimize the backbone configuration for the specific downstream task. The projection head is chosen as a 3 layer MLP with hidden layer dimension of 4096 and output dimension of 128. For data augmentation, random cropping, flipping, color jittering, and Gaussian blur are applied as in \citep{chen2020simple}. For each experiment, the network is pretrained for 1000 epochs on the COCO/2017 \citep{lin2014microsoft} training dataset. The evaluation is performed on the validation subset of COCO/2017 by fixing the pretrained backbone parameters and training a linear classifier on top of the learned representations for multi-label classification. The AdamW \citep{loshchilov2017decoupled} optimizer with a learning rate of $4 \times 10^{-3}$, cosine decay schedule, and a weight decay of $5 \times 10^{-2}$ are used to train the model. To evaluate the performance in multi-label classification evaluations, we report the mAP metric as described in Section 4.2 of \citep{veit2017learning}, and F1-score.

\label{sec:baseline_exp}

\textbf{Baseline experiments} To obtain baseline metrics, we pretrained the same backbone with SimCLR \citep{chen2020simple} and DenseCL \citep{wang2021dense} methods and evaluated the performance on multi-label classification with 
the same parameter setting described in Section \ref{sec:framework}. The dense contrastive loss weight for DenseCL is chosen as $\lambda=0.3$, which performed the best in a linear sweep.

\begin{wrapfigure}{r}{0.45\textwidth}
\centering
\setlength{\tabcolsep}{2.5pt}
\captionof{table}{\small SimCLR and DenseCL multi-label classification results on COCO for different global feature aggregation and dense matching types.}
\begin{tabular}{@{}lcccc@{}}
\toprule
\multicolumn{1}{c}{Method}&\multicolumn{1}{c}{Agg.}&\multicolumn{1}{c}{Pair feature}&\multicolumn{1}{c}{mAP}& \multicolumn{1}{c}{F1}\\
\cmidrule(r){1-1}\cmidrule(lr){2-2}\cmidrule(lr){3-3}\cmidrule(lr){4-4}\cmidrule(lr){5-5}
  {DenseCL} & {CLS} & backbone & \textbf{59.9} & \textbf{38.1} \\
  &  & proj head & 59.8 & \textbf{38.1} \\
  & {GAP} & backbone & 58.1 & 37.3 \\
  &  & proj head & 57.8 & 37.5 \\
  {SimCLR} & CLS & - & \textbf{59.6} & \textbf{37.9} \\
  & GAP & - & 58.4 & 37.7 \\
 \bottomrule
 \vspace{-0.95cm}
\end{tabular}
\label{tab:densecl_metrics}
\end{wrapfigure}

For SimCLR and DenseCL, we test two different methods to obtain global features. The former uses the CLS token of ViT, and the latter uses the average of the dense local features (GAP) as global features. 

For DenseCL, both global and dense projection heads have three linear layers with 4096 hidden dimensionality and global and dense feature vectors are $D=L=128$ dimensional. 
The positive dense pairs are formed based on their cosine similarities using the backbone encoder or global projection head output representations. 
The results are shown in Table \ref{tab:densecl_metrics}.

\begin{wrapfigure}{r}{0.45\textwidth}
\small
\centering
\centering
\includegraphics[scale=0.4]{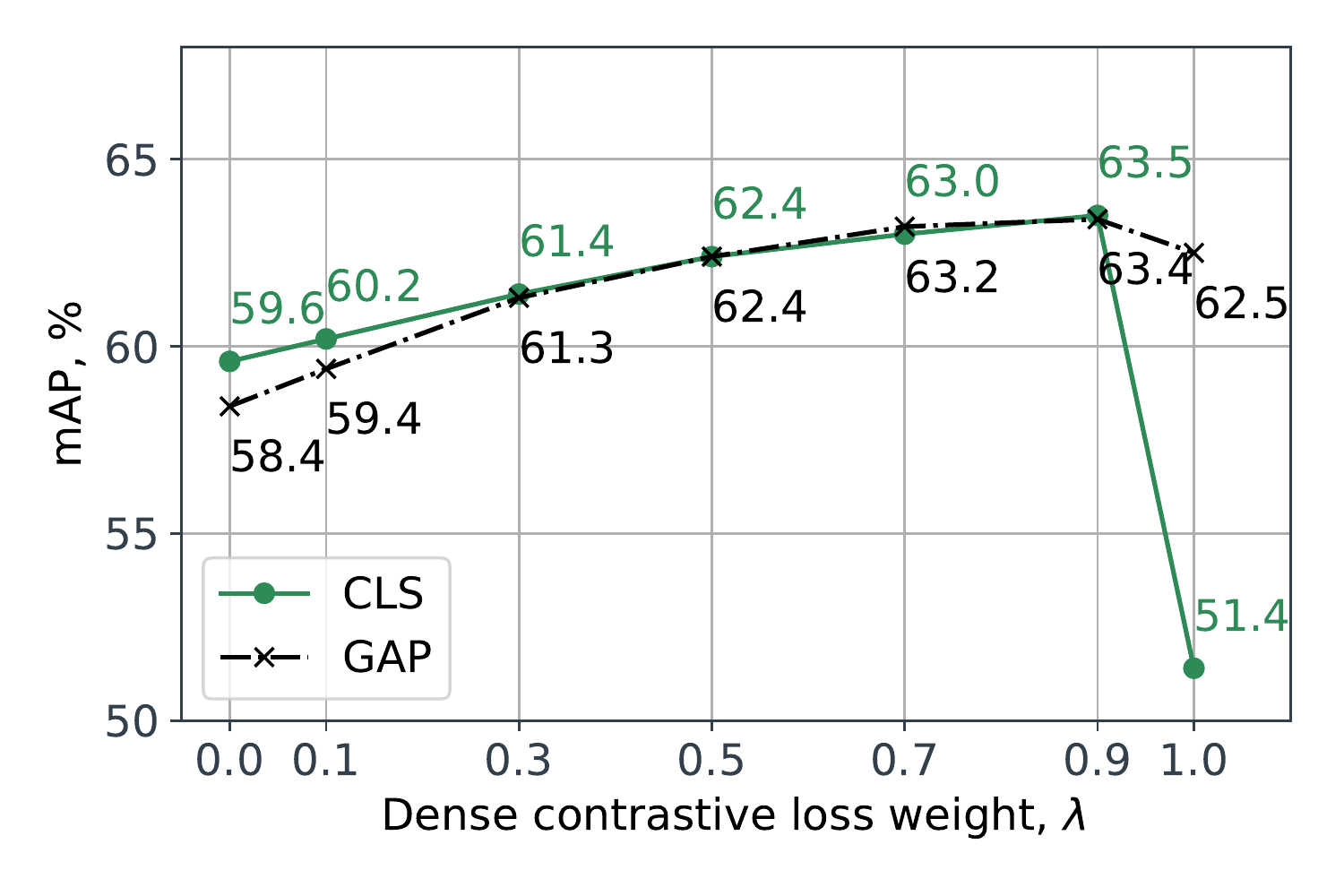}
\captionof{figure}{\small mAP vs. dense contrastive loss weight $\lambda$ plot for DenseCL++ for different global feature aggregation settings.}
\label{fig:rand_dense_neg_sampling}
\end{wrapfigure}

\textbf{DenseCL++ experiments} As mentioned in Section \ref{sec:form_dense_neg_pairs}, we replace dense-global negative comparisons in DenseCL with dense-dense counterparts. 
    A dense feature is sampled uniformly random from each view that belongs to augmentations of different images in the batch to form negative pairs with the dense feature of the anchor view.
    This results in a substantial improvement for evaluation on multi-label classification. The positive pairs are again formed using their 
    encoder output representations. The comparison for different dense loss weights and global feature aggregations 
    techniques 
    as explained in Section \ref{sec:baseline_exp} is reported in Figure \ref{fig:rand_dense_neg_sampling}. The best performing configuration on multi-label classification evaluation 
    is used as the baseline for 
    DenseCL++. 
    We report multi-label classification results for the best performing $M$, $\beta$, and $N$ 
    in Table \ref{tab:densecl_metrics1}.  
    In Table \ref{tab:voc_coco_seg}
    we show that DenseCL++ can also lead to improvements when fine-tuning models for semantic segmentation on PASCAL VOC and MS COCO. Experimental details for segmentation evaluation can be found in Appendix~\ref{app:segmentation}. We also provide further ablation studies with various dense negative formulation techniques and reconstruction modules in Appendix~\ref{app:ablations}.
    
    The results in Figure \ref{fig:rand_dense_neg_sampling} show that the performance of DenseCL++ consistently improves until $\lambda=0.9$ and degrades for $\lambda=1$, where the drop is more dramatic for the CLS aggregation. Both CLS and GAP cases work almost equally well for the optimal weight of $\lambda=0.9$ but due to the significant drop at $\lambda=1$ for CLS aggregation, we report GAP aggregation results for DenseCL++ in Table \ref{tab:densecl_metrics1}.

    \begin{figure}[hbt]
    \small
    \vspace{0.25cm}
    \begin{minipage}{0.45\textwidth}
    \captionof{table}{\small Top 
    performances
    for multi-label classification on COCO 
    for
    different 
    methods. Baseline DenseCL++ forms dense negative pairs as in Fig.~\ref{fig:densecl_dense_neg_diagrams}(a) whereas DenseCL++* uses guided dense negatives with $M\text{=}256$, $\beta\text{=}0.5$ and $N\text{=}64$.}
    \setlength{\tabcolsep}{1.2pt}
    \centering
    \begin{tabular}{@{}lcccc@{}}
    \toprule
    \multicolumn{1}{c}{Method}&\multicolumn{1}{c}{Agg.}&\multicolumn{1}{c}{Pair feature}&\multicolumn{1}{c}{mAP}& \multicolumn{1}{c}{F1}\\
    \cmidrule(r){1-1}\cmidrule(lr){2-2}\cmidrule(lr){3-3}\cmidrule(lr){4-4}\cmidrule(lr){5-5}
      {SimCLR} & {CLS} & $-$ & 59.6 & 37.8 \\
      {DenseCL} & {CLS} & backbone & 59.9 & 38.1 \\
      {DenseCL++} & {GAP} & backbone & \textbf{63.4} & \textbf{39.0} \\ 
      {DenseCL++*} & {GAP} & backbone & \textbf{64.1} & \textbf{39.1} \\
     \bottomrule
     \vspace{-0.45cm}
    \end{tabular}
    \label{tab:densecl_metrics1}
    \end{minipage}
\hspace{0.5cm}
\begin{minipage}{0.5\textwidth}
\centering
\captionof{table}{\small Semantic segmentation on PASCAL VOC and MS COCO. All methods use GAP aggregation type. Experimental details are provided in Appendix \ref{app:segmentation}.}
\setlength{\tabcolsep}{1pt}
\begin{tabular}{@{}lccc@{}}
\toprule
\multicolumn{1}{c}{Method}&\multicolumn{1}{c}{Pair feature}&\multicolumn{1}{c}{VOC mIoU} & \multicolumn{1}{c}{COCO mIoU} \\
\cmidrule(r){1-1}\cmidrule(lr){2-2}\cmidrule(lr){3-3}\cmidrule(lr){4-4}
  SimCLR & $-$ & 69.3 & 61.5 \\
  DenseCL++ & backbone & \textbf{70.0} & \textbf{63.3}\\
 \bottomrule
\end{tabular}
\label{tab:voc_coco_seg}
\end{minipage}
\vspace{0.25cm}
\end{figure}

\textbf{Acknowledgement} We thank our colleagues from Google Research and Brain, Dilip Krishnan, Yin Cui, Aaron Sarna, Ting Chen, and Golnaz Ghiasi who provided insight and expertise that greatly assisted this research. Also, we are grateful to Yeqing Li who provided the UViT implementation.

{
\small
\bibliographystyle{abbrvnat}
\bibliography{egbib}
}

\appendix
\part*{Appendix}

\section{PASCAL VOC and COCO Semantic Segmentation}
\label{app:segmentation}
To further evaluate the downstream performance of the proposed pretraining scheme, we finetune our models with a UViT \citep{chen2021simple} backbone on PASCAL VOC and COCO/2014 semantic segmentation tasks. As in the multi-label classification case, we do not explicitly try to optimize the backbone configuration. Thus, the UViT architecture configuration is the same across different methods and has depth of 18 layers, hidden size of 342, number of attention heads per layer as 6, and patch size of~8. 
We also do not leverage the specific attention window strategy of UViT and use global attentions. DenseCL++ forms dense negative pairs using the random sampling method described in Fig. \ref{fig:densecl_dense_neg_diagrams} (a) and has a dense contrastive loss weight $\lambda = 0.7$.

All segmentation models are finetuned using a COCO pretrained initialization as in multi-label classification experiments. 
PASCAL VOC segmentation evaluations use ASPP \citep{chen2017deeplab} decoder, DeepLabv3+ \citep{chen2018encoder} head, AdamW \citep{loshchilov2017decoupled} optimizer with a stepwise decay schedule with initial learning rate of $3\times 10^{-3}$, weight decay as $10^{-4}$, batch size of 256, and 20k training steps. COCO segmentation evaluations use FPN \citep{lin2017feature} decoder, AdamW optimizer with initial learning rate of $5 \times 10^{-4}$ with cosine decay schedule, batch size of 256, and 64k training steps.
The results are shown in Table \ref{tab:voc_coco_seg}. 

\section{Reconstruction Decoder}

\begin{figure}[ht!]
\centering
\includegraphics[scale=0.7,trim={0 20pt 50pt 20pt},clip]{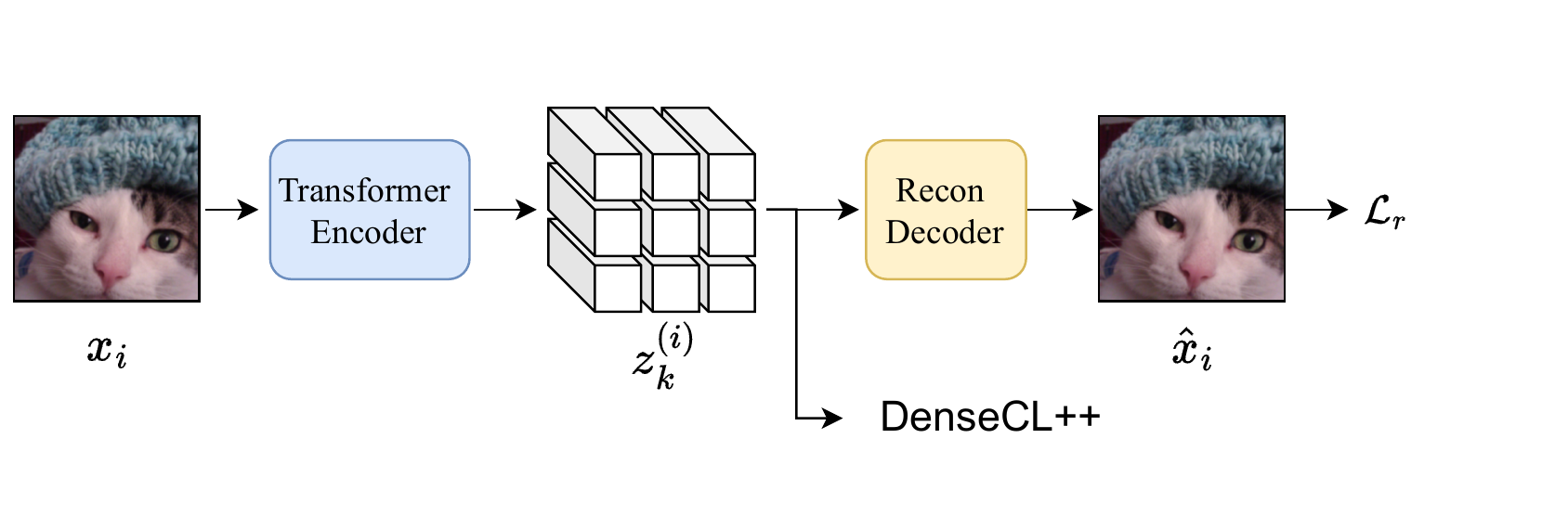}
\caption{\small Reconstruction module for DenseCL++ framework.}
\label{fig:densecl_dense_recon}
\end{figure}

Inspired by several recent studies that incorporate an auxiliary lightweight reconstruction module in the contrastive learning framework \citep{li2021mst, wang2022repre}, we test several different alternatives in the context of our proposed dense contrastive loss. These include convolutional and transformer-based decoders that reconstruct the augmented input image from its dense hidden representations. The reconstruction architecture is shown in Fig. \ref{fig:densecl_dense_recon}. 

\subsection{Reconstruction Objective}
\label{app:reconstruction}

The encoder and the reconstruction network parameters are updated end-to-end during pre-training to minimize the mean absolute error,
\begin{equation}
    \label{eq:recon_loss}
    \mathcal{L}_{r} = ||x_i - \hat{x}_i||,
\end{equation}
where $\hat{x}_i$ is the estimated reconstruction of the input image. The reconstruction loss term is weighted by a constant $\gamma \in \mathbb{R}$. The overall loss is expressed as
\begin{equation}
    \mathcal{L} = (1-\lambda)\mathcal{L}_{g} + \lambda\mathcal{L}_{d} + \gamma\mathcal{L}_{r}.
\end{equation}
The decoder is discarded after pre-training of the encoder. 

\subsection{Convolutional Decoder}
In this setting a simple convolutional neural network architecture with consecutive convolutional layers is applied on the global or dense features of the input image. Global features are reshaped to 2D before they are fed to the decoder. On the other hand, dense representations $z_k^{(i)} \in \mathbb{R}^{S \times S \times L}$ are treated as $L$ channel inputs. In both cases, either transposed convolutional layers or convolutional layers with fixed number of output channels and a fixed rate of bicubic upsampling are applied consecutively to the inputs at each layer until the estimated reconstruction $\hat{x} \in \mathbb{R}^{224 \times 224 \times 3}$ is obtained. 

\subsection{Transformer-based Decoder}
It is also possible to use the transformer architecture \citep{dosovitskiy2020image} to reconstruct images. To do so, a linear layer is applied to the dense features to project them to a lower dimensional subspace. Then, the projected representations are fed to the decoder and the image is reconstructed patch-wise as in \citep{he2022masked}.

\section{Ablation studies}
\label{app:ablations}

\subsection{Dense negative pair forming strategies}

\begin{figure}[hbt]
    \centering
    \begin{subfigure}{.32\textwidth}
    \centering
    \includegraphics[width=\linewidth]{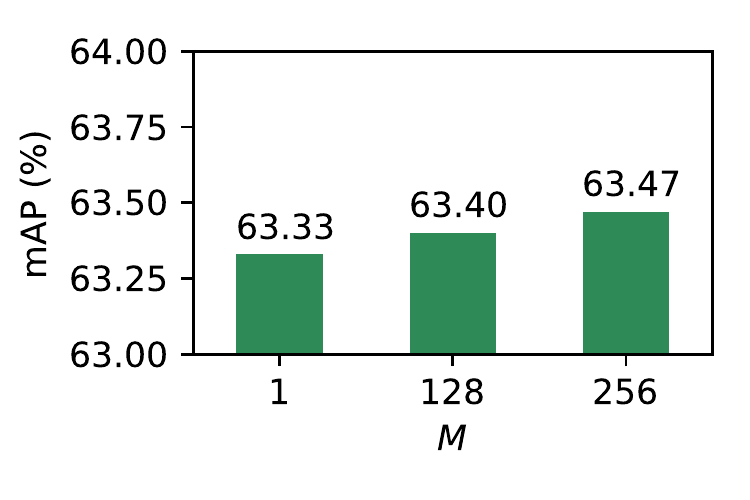}
    \caption{}
    \end{subfigure} 
    \begin{subfigure}{.32\textwidth}
    \centering
    \includegraphics[width=\linewidth]{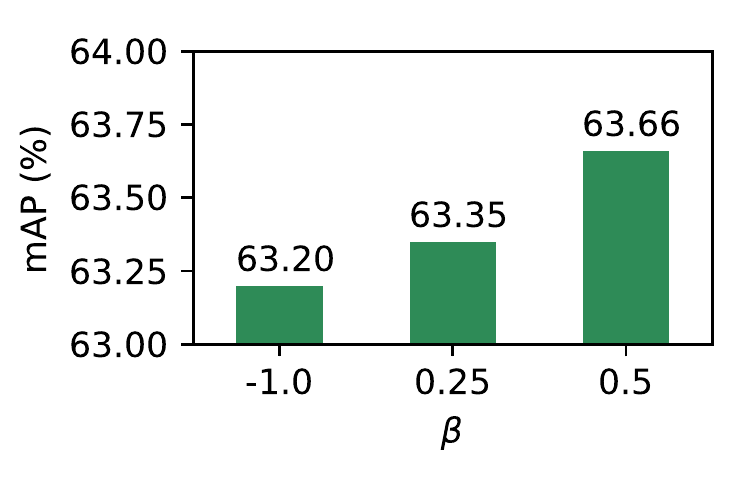}
    \caption{}
    \end{subfigure} 
    \begin{subfigure}{.32\textwidth}
    \centering
    \includegraphics[width=\linewidth]{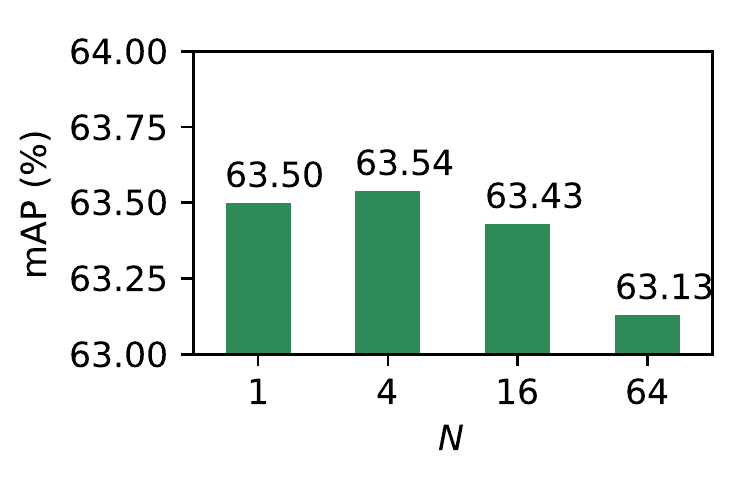}
    \caption{}
    \end{subfigure}
    \caption{\small Averaged mAP for each fixed parameter. We conduct 36 experiments with all different configurations of $M$, $\beta$, $N$ and report the averaged metrics for experiments that include the specific parameter value. Results for different (a) number of random negative index sampling sets, (b) similarity thresholds, and (c) cross-view negative pairs.}
    \label{fig:avg_M_thr_N_sweep}
\end{figure}

    \textbf{Guided dense negative sampling from other pairs.} Encouraged by the performance improvement introduced by random dense negative sampling, we also experiment with a guided dense negative sampling method. We repeat the random sampling of dense features from other pairs $M$ times and pick the set of dense features that have the largest average similarity to the anchor view features. 
    As described before, we aim to form hard dense negatives as a result of this modification. Averaged mAP values over $\beta$ and $N$ for different $M$ are shown in Fig. \ref{fig:avg_M_thr_N_sweep} (a).

    Furthermore, as described in Section \ref{sec:form_dense_neg_pairs}, hard thresholding can be applied to the similarities while computing the most similar set of dense negative features on average to the anchor image features. By doing so, it can be prevented that the average similarity being mostly decided by moderately similar dense features of the anchor image which ultimately assigns a more important role to hard dense negatives. Fig. \ref{fig:avg_M_thr_N_sweep} (b) shows the mAP performance for various threshold levels $\beta$. 

\begin{figure}[hbt]
    \centering
    \begin{subfigure}{.22\textwidth}
    \centering
    \includegraphics[width=\linewidth]{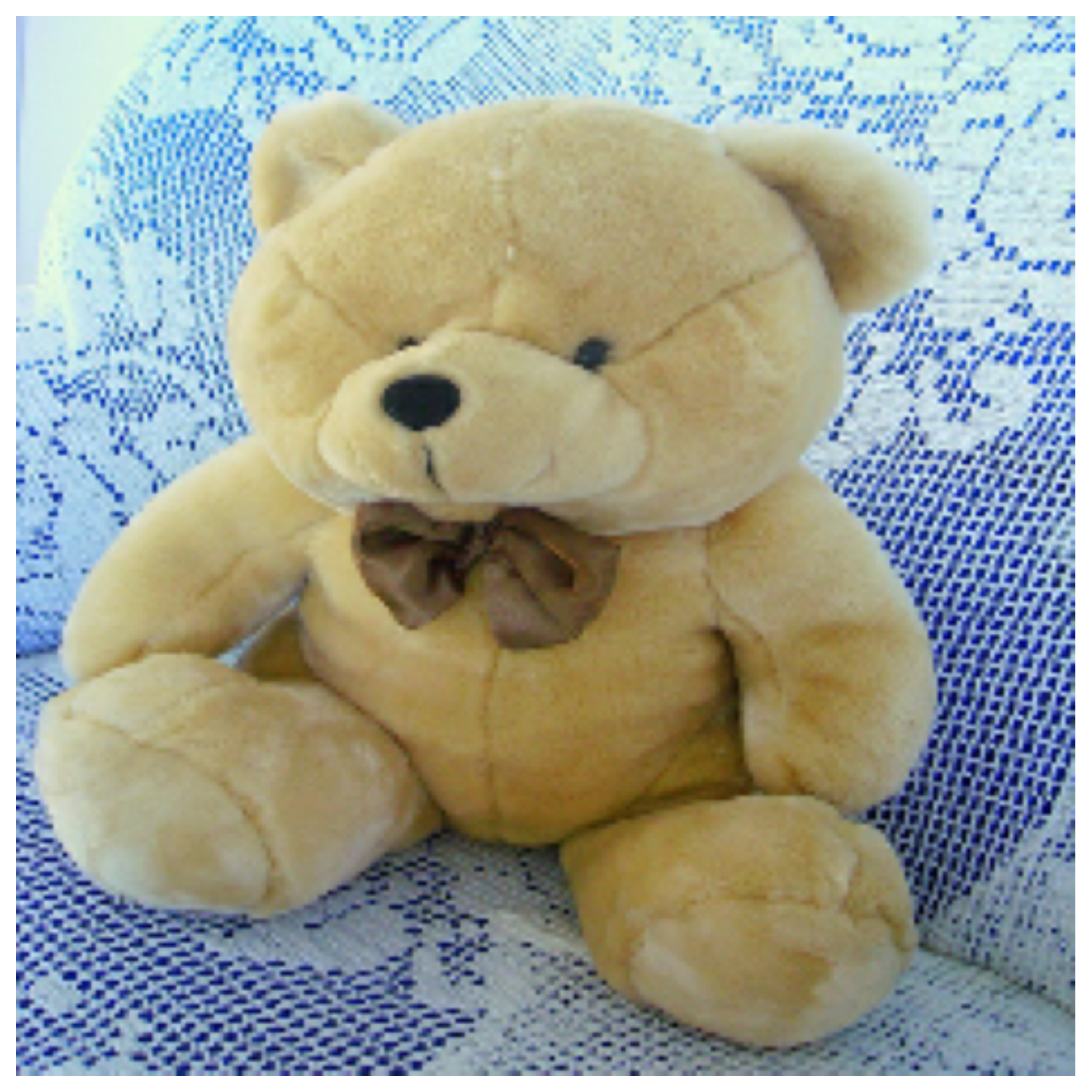}
    \caption{}
    \end{subfigure} 
    \begin{subfigure}{.22\textwidth}
    \includegraphics[width=\linewidth]{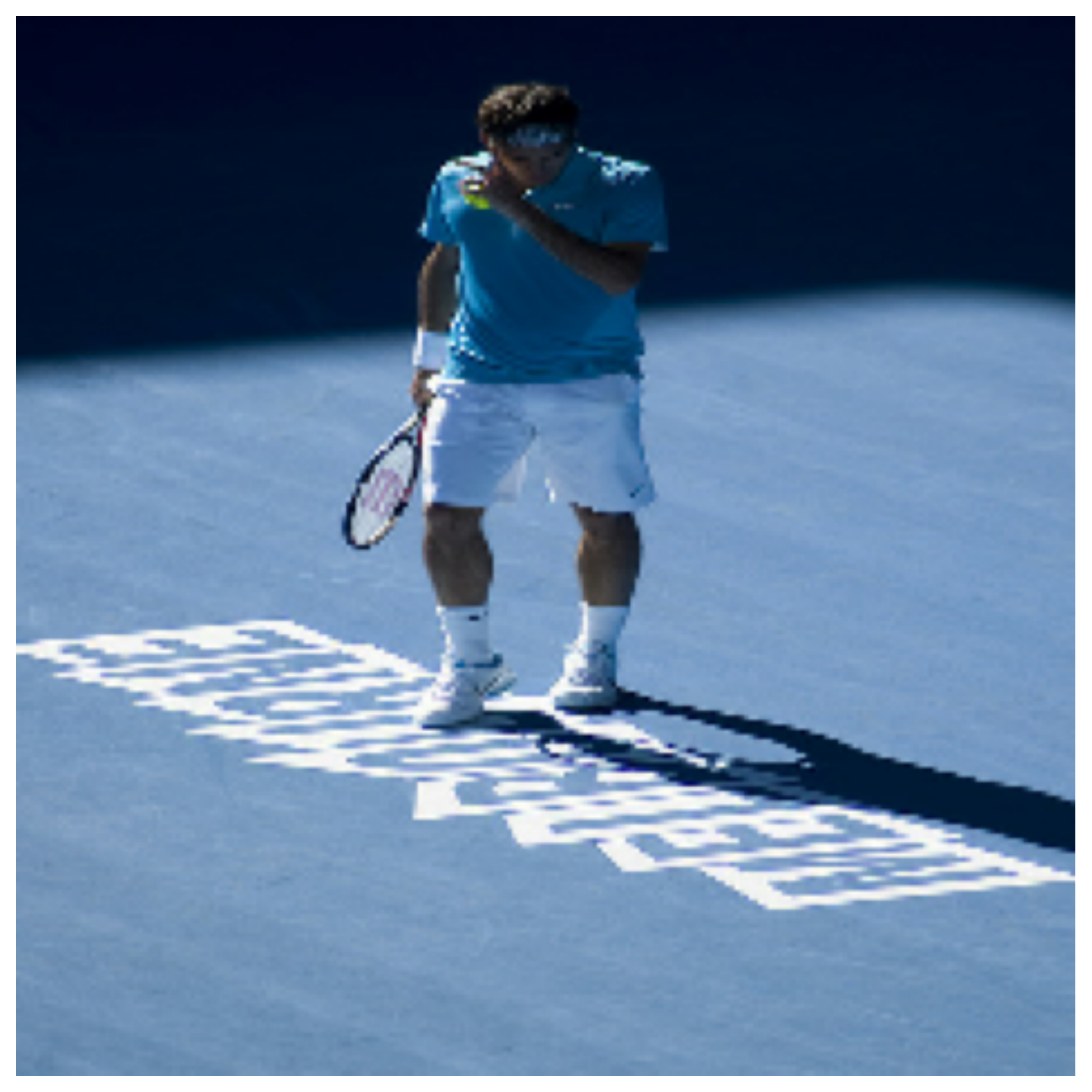}
    \caption{}
    \end{subfigure} \\
    \begin{subfigure}{.27\textwidth}
    \includegraphics[width=\linewidth]{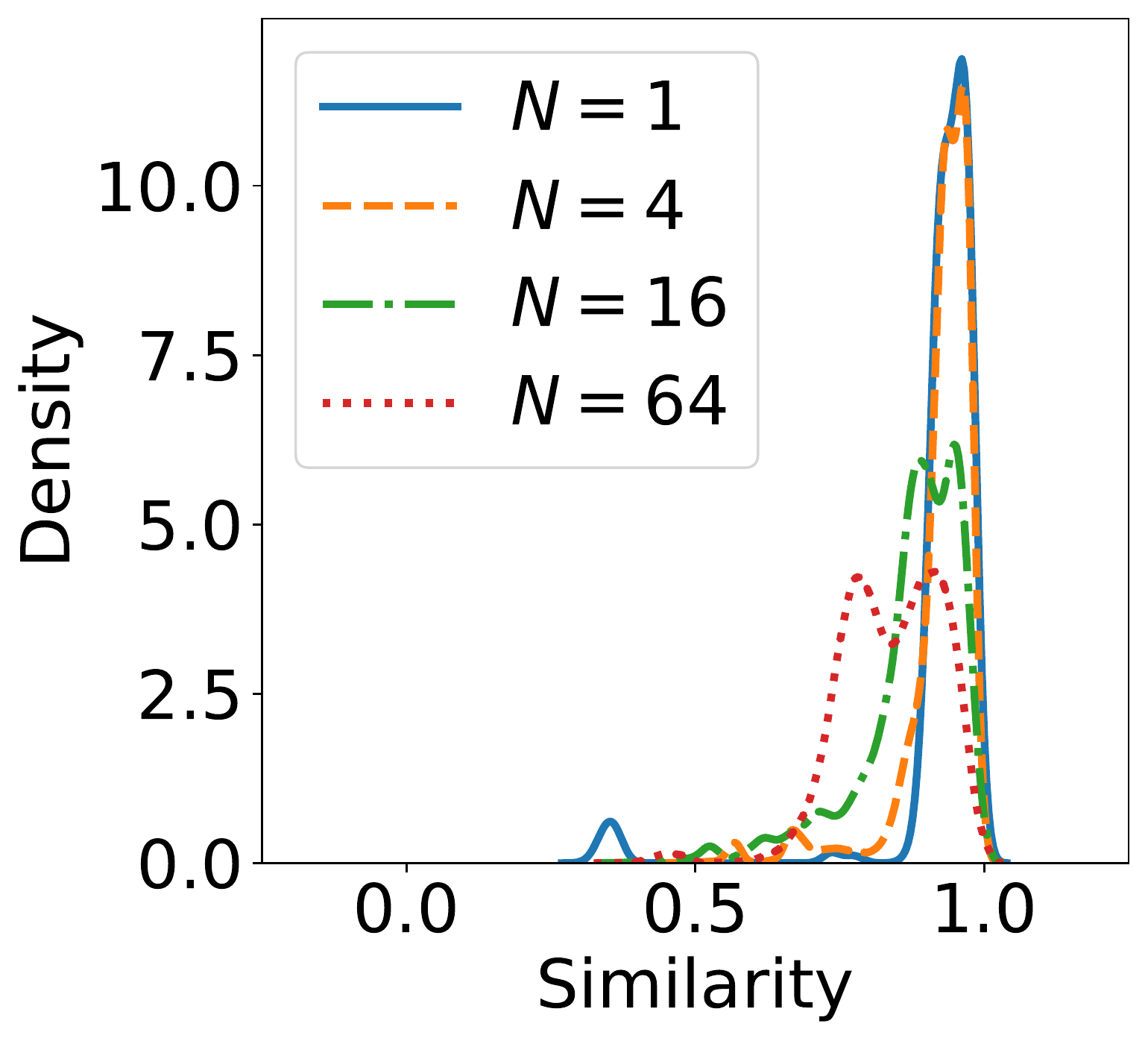}
    \caption{}
    \end{subfigure}
    \begin{subfigure}{.27\textwidth}
    \includegraphics[width=\linewidth]{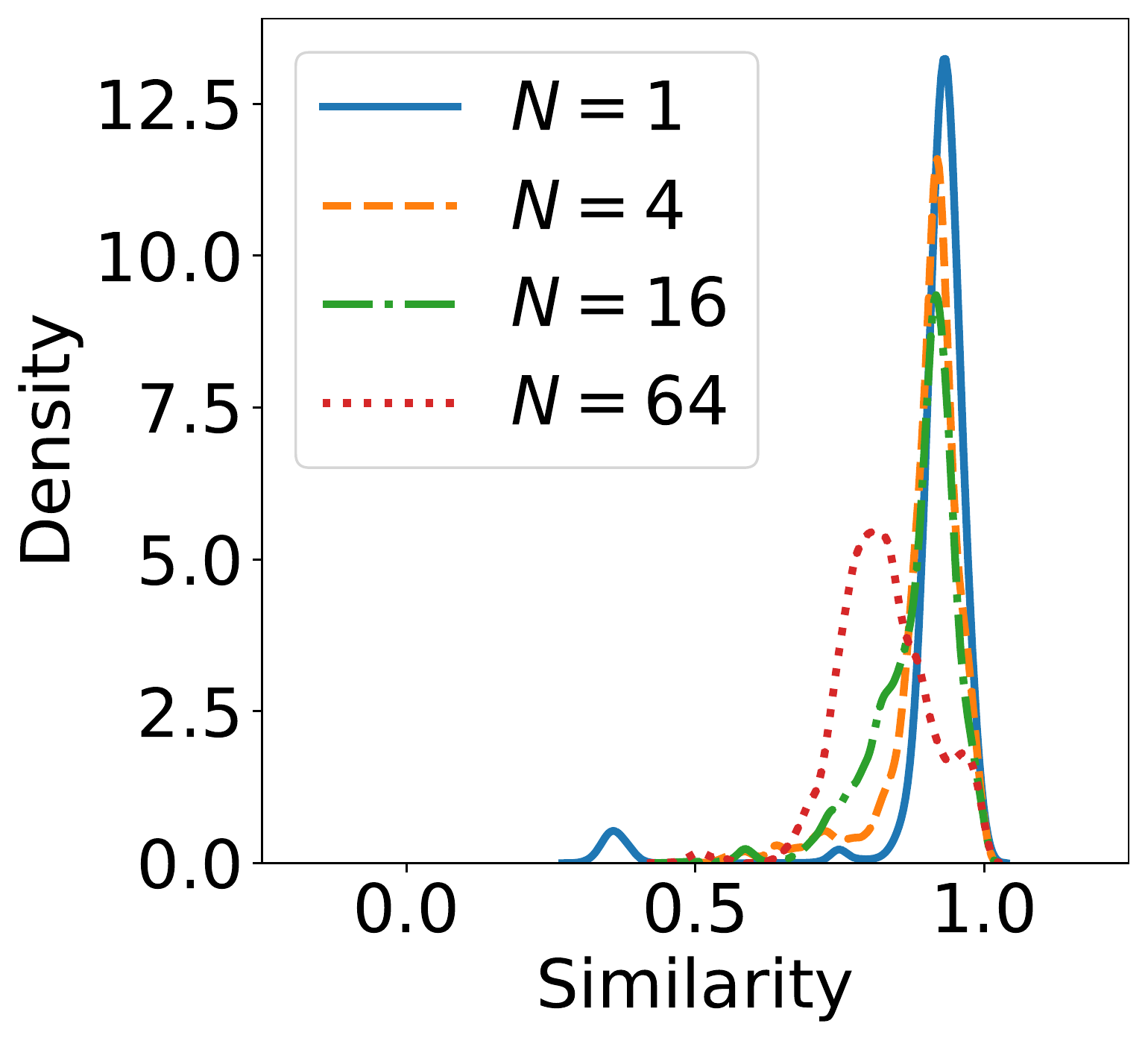}
    \caption{}
    \end{subfigure}
    \begin{subfigure}{.255\textwidth}
    \includegraphics[width=\linewidth]{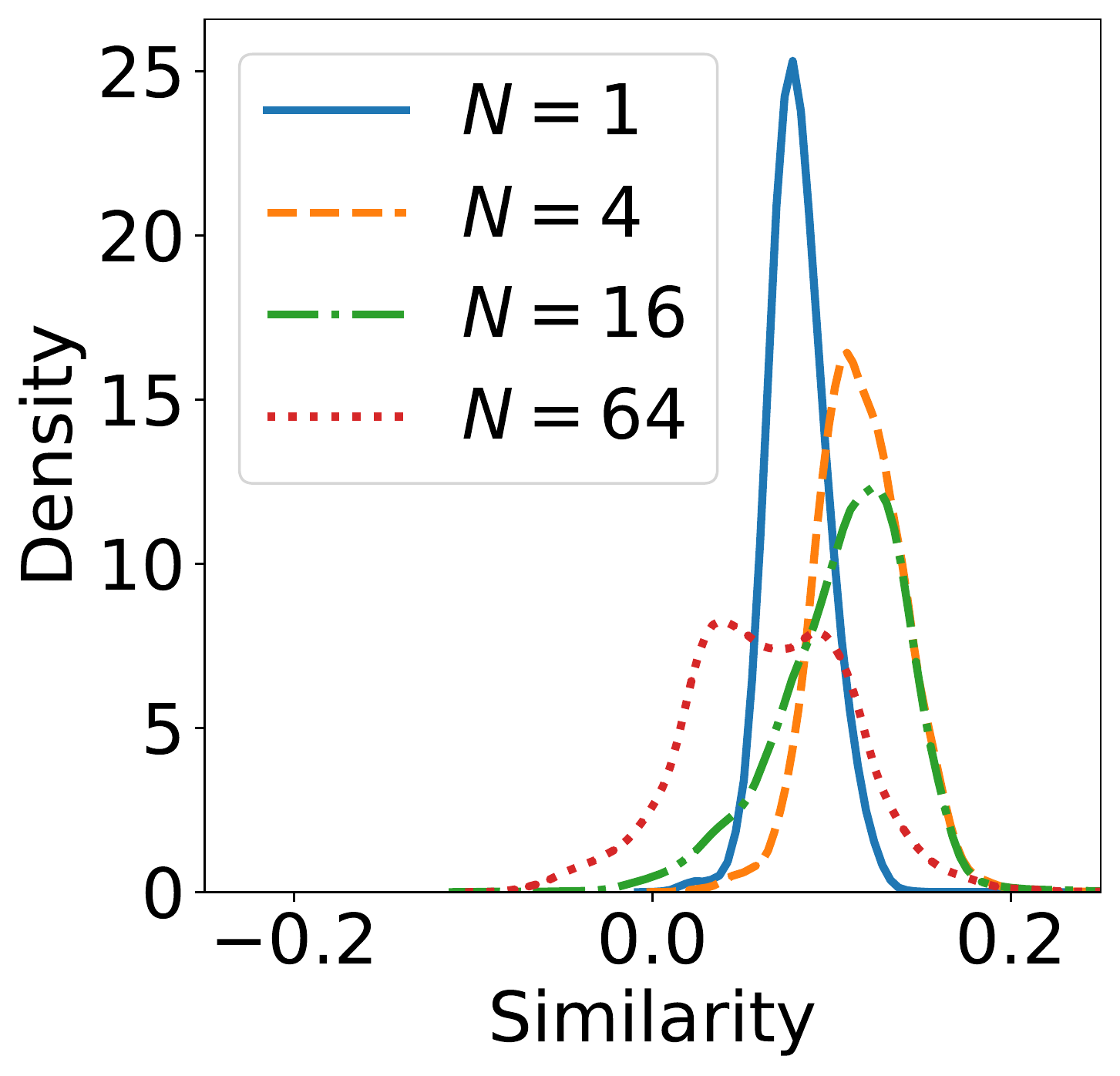}
    \caption{}
    \end{subfigure}
    \caption{\small Intra-image and inter-image dense feature cosine similarity distributions of images (a) and (b) for increasing number of cross-view dense negatives $N$ during training. Graphs (c) and (d) show the intra-image distributions for (a) and (b), respectively. Inter-image distributions are shown in (e).}
    \label{fig:similarity_hist_comp}
\end{figure}

    \textbf{Sampling dense negatives from the corresponding view.}
    After observing high and condensed similarity distributions for the augmented views of the same image during training and evaluation, we also compute pairwise dense negatives by finding the least similar $k$ dense features in the corresponding view of the anchor dense feature map. As shown in Fig. \ref{fig:similarity_hist_comp}, incorporating additional pairwise dense negatives results in widened cross similarity distributions. Evaluation performances are provided in Fig. \ref{fig:avg_M_thr_N_sweep} (c).

\subsection{Dense positive pair forming strategies}

\begin{figure}
    \centering
    \includegraphics[scale=0.38]{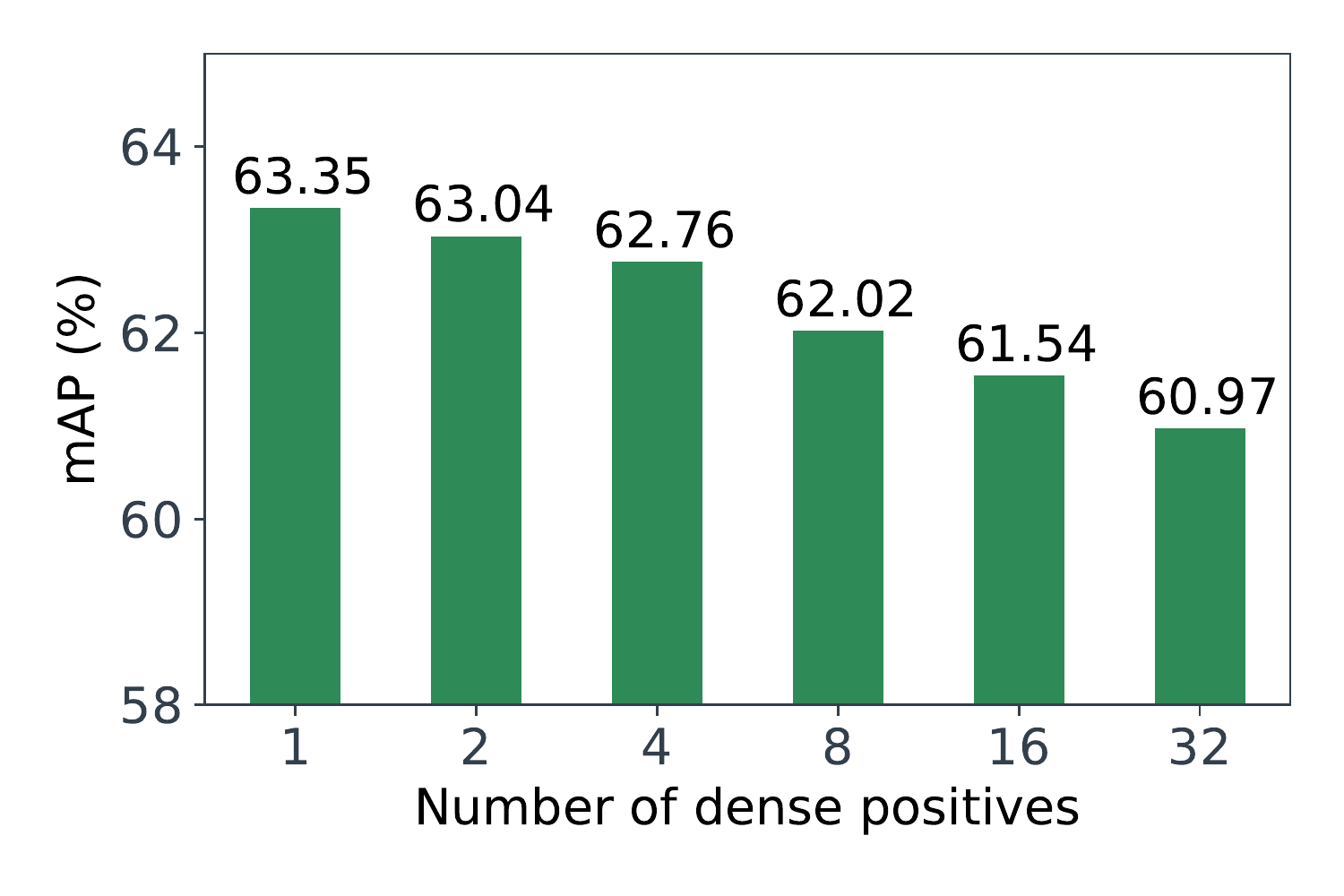}
    \caption{\small mAP vs. number of cross view positive pairs. The rest of the hyperparameters pre-training are selected identically to the baseline DenseCL++ configuration described in Table \ref{tab:densecl_metrics1}. Larger number of cross view dense positives leads to decreasing evaluation performances consistently.}
    \label{fig:densecl_guided_dense_pos}
    \end{figure}

    To explore the potential effect of introducing multiple positive correspondences in our framework, we experiment with multiple positive pair sampling from the corresponding view by selecting the top-$k$ similarity pairs as positive correspondences using the best performing DenseCL++ configuration with random dense negative sampling from other view pairs. We adapt our loss function provided in \eqref{eq:densecl_w_dense_neg_pairs} such that it incorporates multiple positive dense features as proposed in \citep{khosla2020supervised}. The performances for increasing number of positives are provided in Fig. \ref{fig:densecl_guided_dense_pos}.
    
\subsection{Reconstruction module strategies}
\label{sec:recon_module_str}

        \textbf{Convolutional decoder.} Using reshaped global features as input to the convolutional reconstruction decoder with various projected representation dimensions do not provide interpretable reconstructions and hence do not have a considerable effect on the evaluation performance of the method. Thus, we report the experimental results belonging to dense feature input setting. Although incorporation of the dense convolutional decoder do not improve multi-label classification evaluation performance considerably over our DenseCL++, we observe that convolutional layers followed by a fixed interpolation scheme, e.g. bicubic, works slightly better compared to using transposed convolutional layers. Also, increasing the loss weight above a certain level and hence achieving better reconstruction estimates degrade the evaluation performance significantly.
        
        The best performing configuration in our experiments provided 63.6$\%$ mAP and 38.9$\%$ F1-score, approximately +0.2$\%$ mAP improvement over the baseline DenseCL++ performance. In this setting, we used $\gamma = 1 \times 10^{-6}$, number of output layers per convolutional layer as 16, and 4x bicubic upsampling after each convolutional layer.

\begin{figure}[hbt]
    \centering
    \begin{subfigure}{.3\textwidth}
    \centering
    \includegraphics[width=\linewidth]{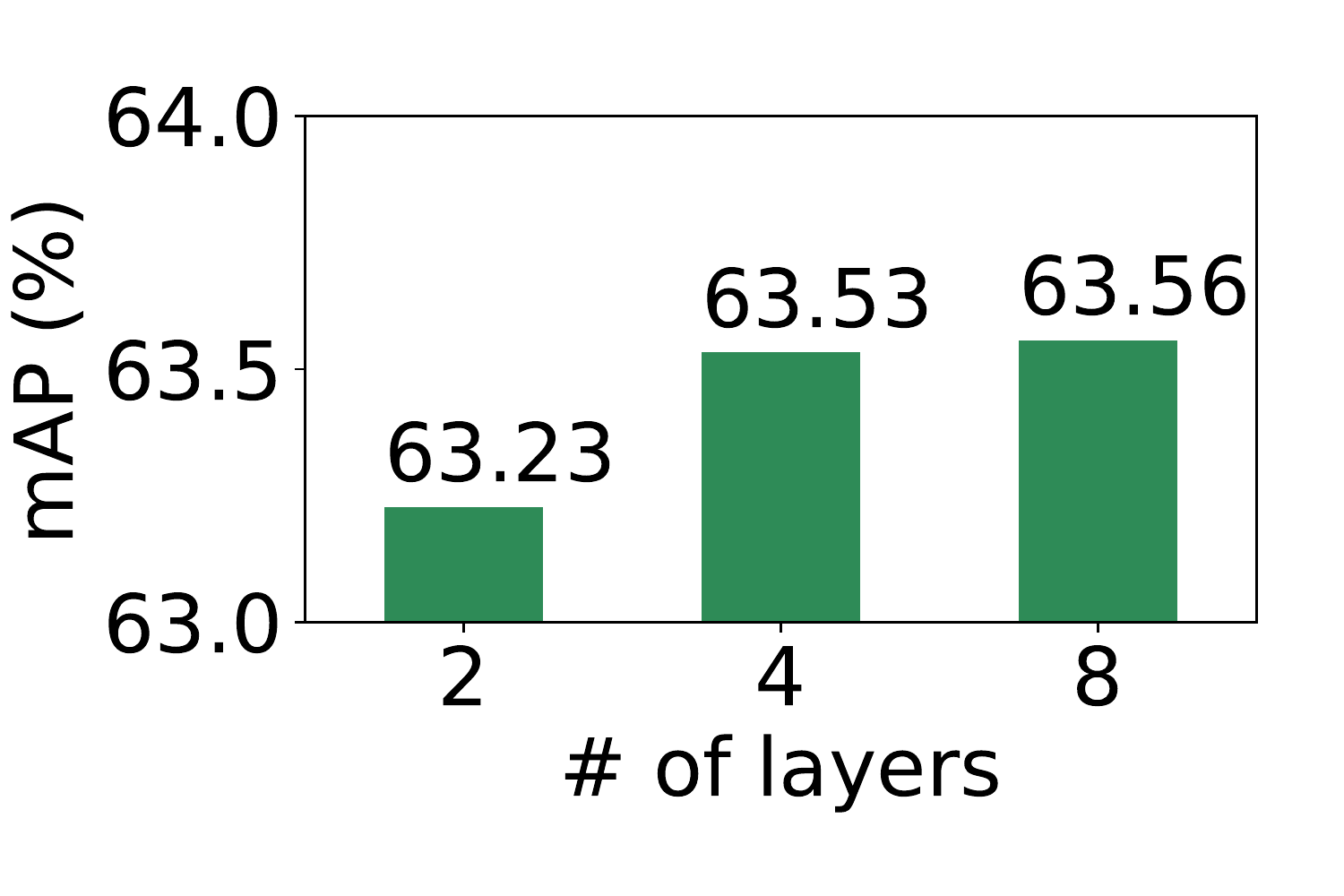}
    \caption{}
    \end{subfigure} 
    \begin{subfigure}{.3\textwidth}
    \centering
    \includegraphics[width=\linewidth]{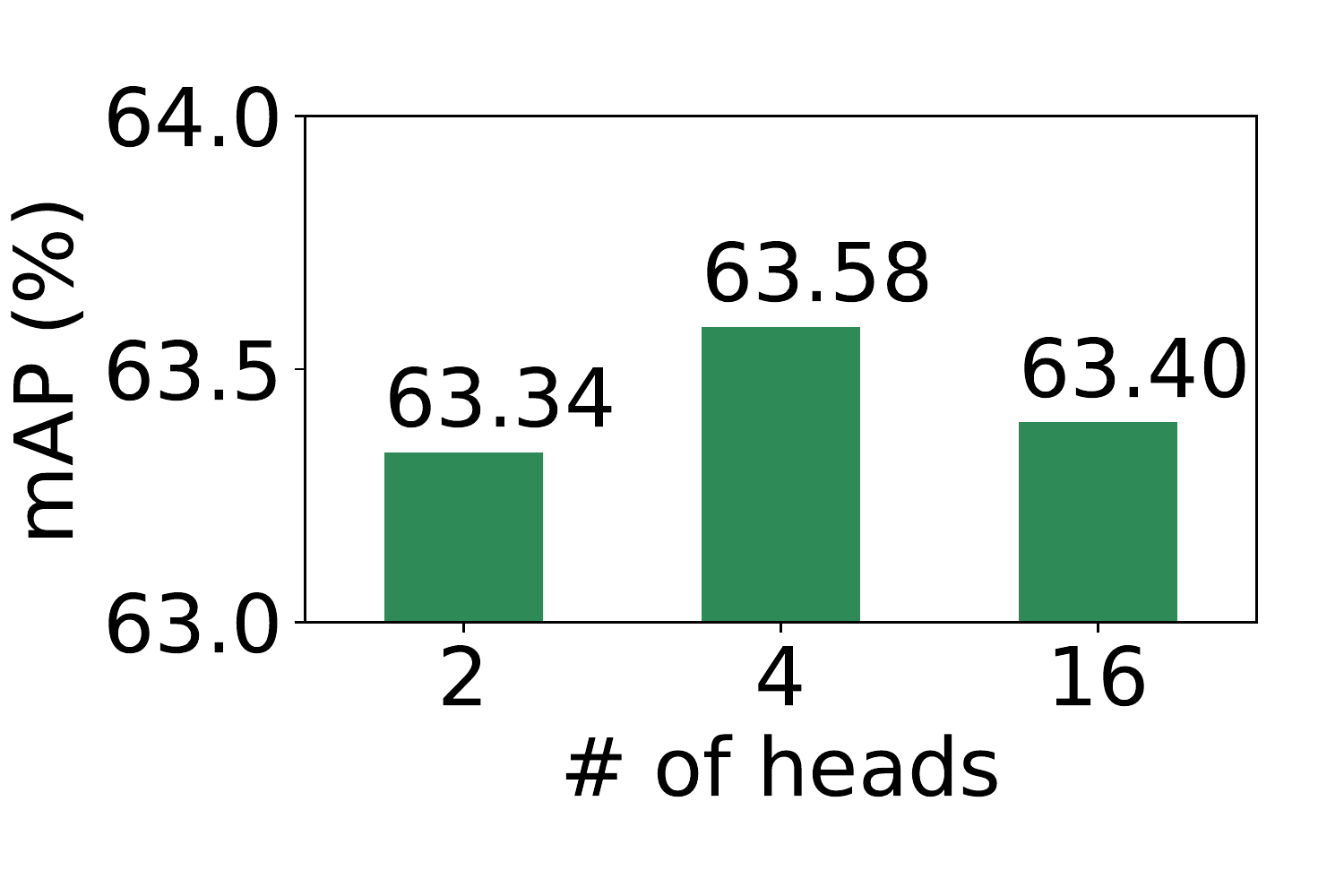}
    \caption{}
    \end{subfigure}
    \caption{\small Averaged mAP for fixed levels of number of (a) layers and (b) heads in the transformer-based decoder for decoder hidden dimension of 64. In total, 9 different experiments were conducted with all possible configurations for number of heads and layers in the decoder and the averaged mAP for experiments with the specific parameter value are reported. Remaining pre-training hyperparameters were kept the same as in baseline DenseCL++ in Table \ref{tab:densecl_metrics1}.
    }
    \label{fig:transformer_heads_layers}
    \end{figure}

        \textbf{Transformer-based decoder.} Similar to the convolutional decoder setting we did not observe considerable improvement for multi-label classification evaluation performance when transformer-based decoder is used. Again, prioritizing reconstruction performance causes a performance drop in the evaluation stage. Both CLS token and GAP aggregation settings for obtaining global hidden representations perform similarly. Reasonable reconstructions can be achieved for projected decoder hidden representations with dimension 32 or larger. Lower dimensional vectors cause significant block-like artifacts. In accordance with the suggestions of using a lightweight decoder in \citep{li2021mst} and \citep{wang2022repre}, we also experiment with reduced number of transformer blocks and attention heads for each block compared to \citep{he2022masked} and provide results in Figure \ref{fig:transformer_heads_layers}.
        
        Using transformer-based decoder provides a marginal improvement (+0.3$\%$ mAP) over the baseline DenseCL++ result in the best case when number of layers and number of heads are both 4 and the loss weight is $\gamma = 5 \times 10^{-7}$ with decoder latent dimension of 64.
        
\begin{figure}[hbt]
        \centering
        \includegraphics[width=0.5\linewidth]{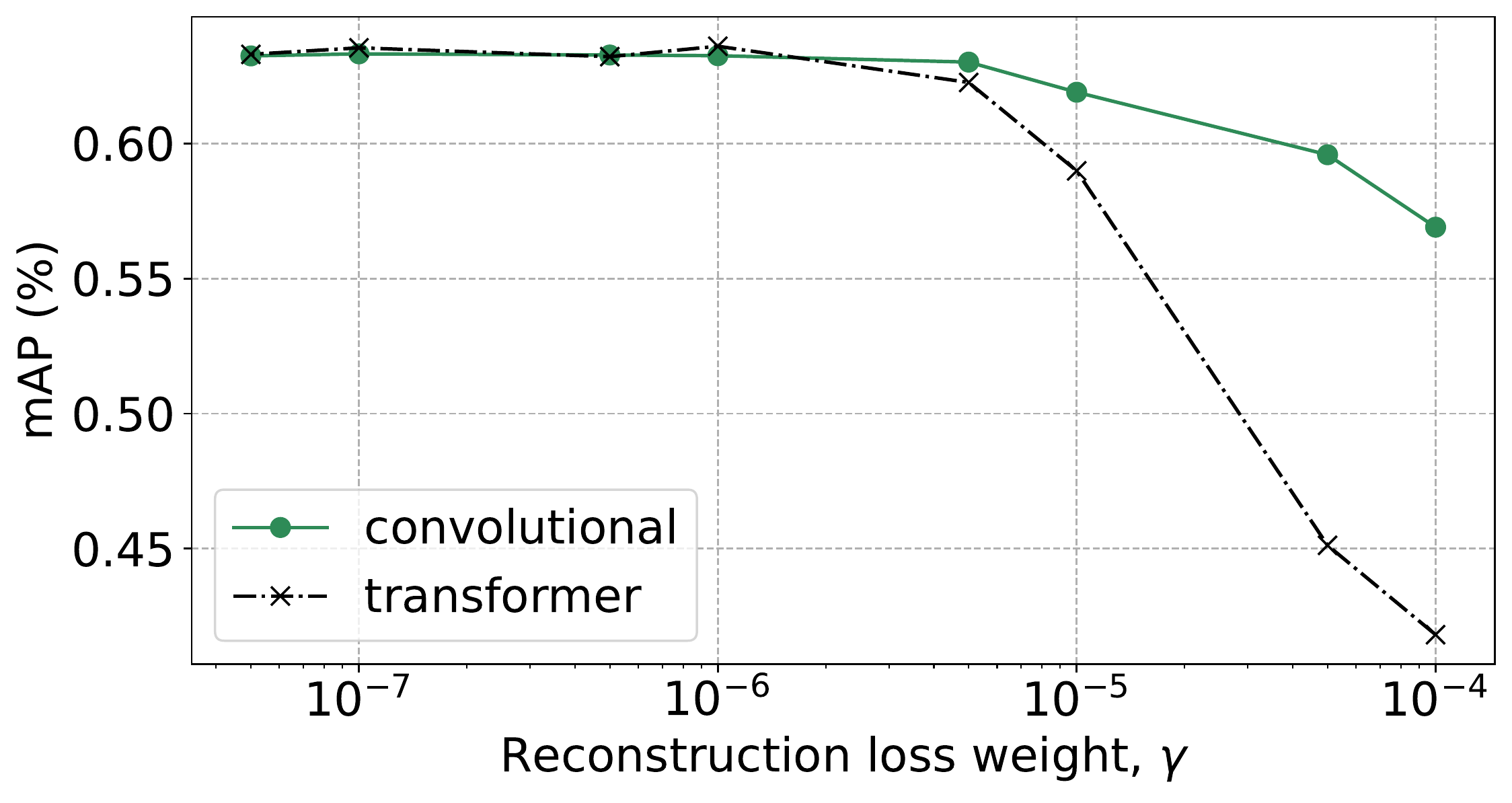}
        \caption{\small The multi-label classification mAP vs. reconstruction loss weight $\gamma$ for convolutional and transformer-based decoders. Convolutional decoder has the configuration described in Section \ref{sec:recon_module_str} and the transformer-based uses 16 layers and 8 attention heads at each layer.}
        \label{fig:recon_loss_sweep}
    \end{figure}

\begin{figure}[hbt]
        \centering
        \begin{subfigure}{.18\textwidth}
        \centering
        \includegraphics[width=0.99\linewidth]{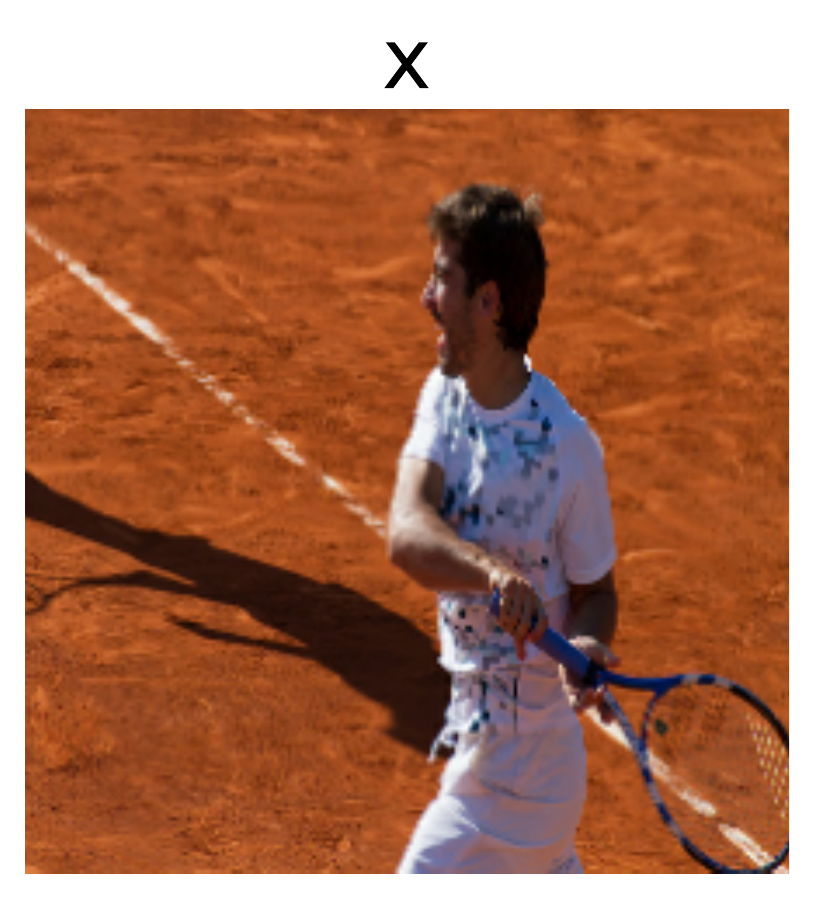}
        \caption{Input}
        \end{subfigure} 
        \begin{subfigure}{.18\textwidth}
        \includegraphics[width=0.99\linewidth]{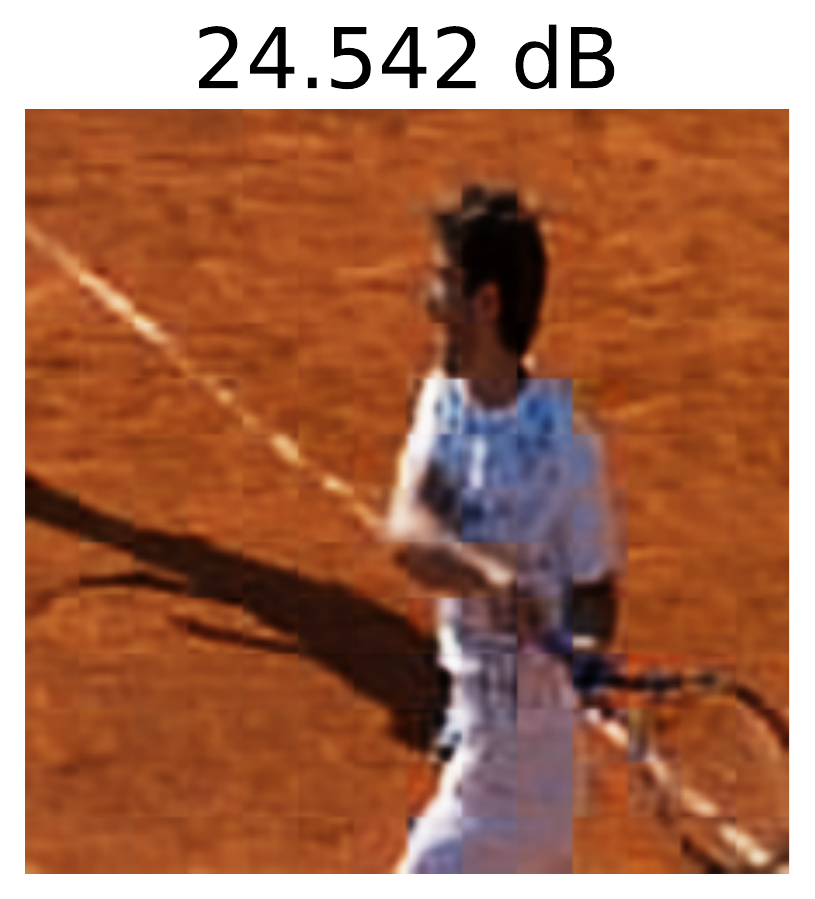}
        \caption{$\gamma = 10^{-5}$}
        \end{subfigure}
        \begin{subfigure}{.18\textwidth}
        \includegraphics[width=0.99\linewidth]{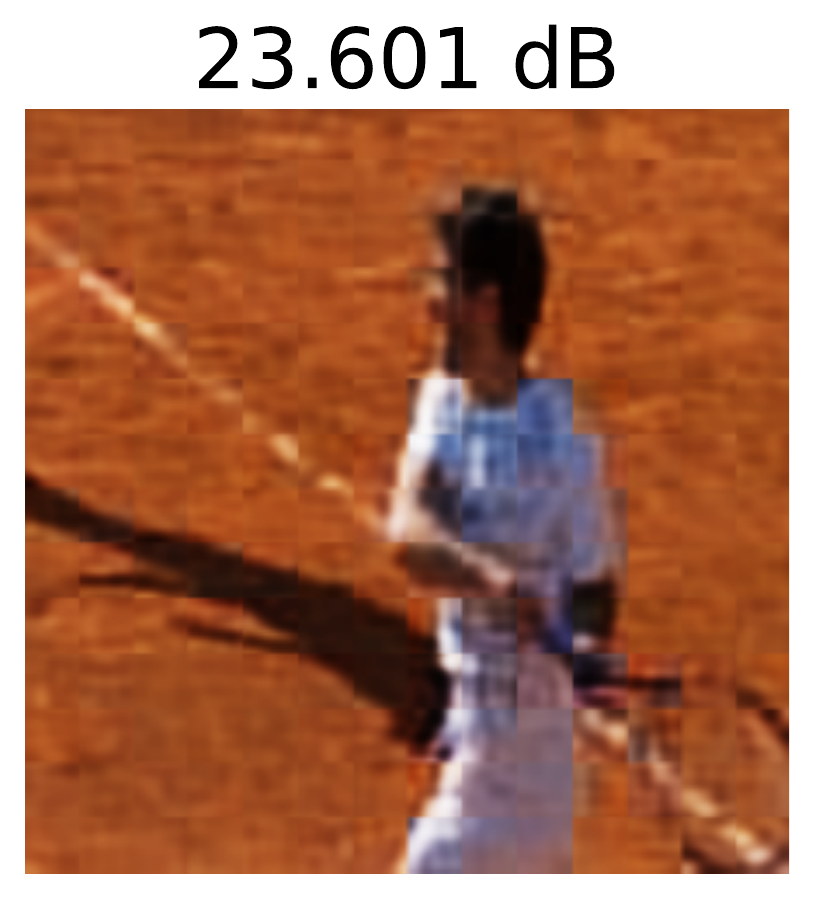}
        \caption{$\gamma = 5 \times 10^{-6}$}
        \end{subfigure}
        \begin{subfigure}{.18\textwidth}
        \includegraphics[width=0.99\linewidth]{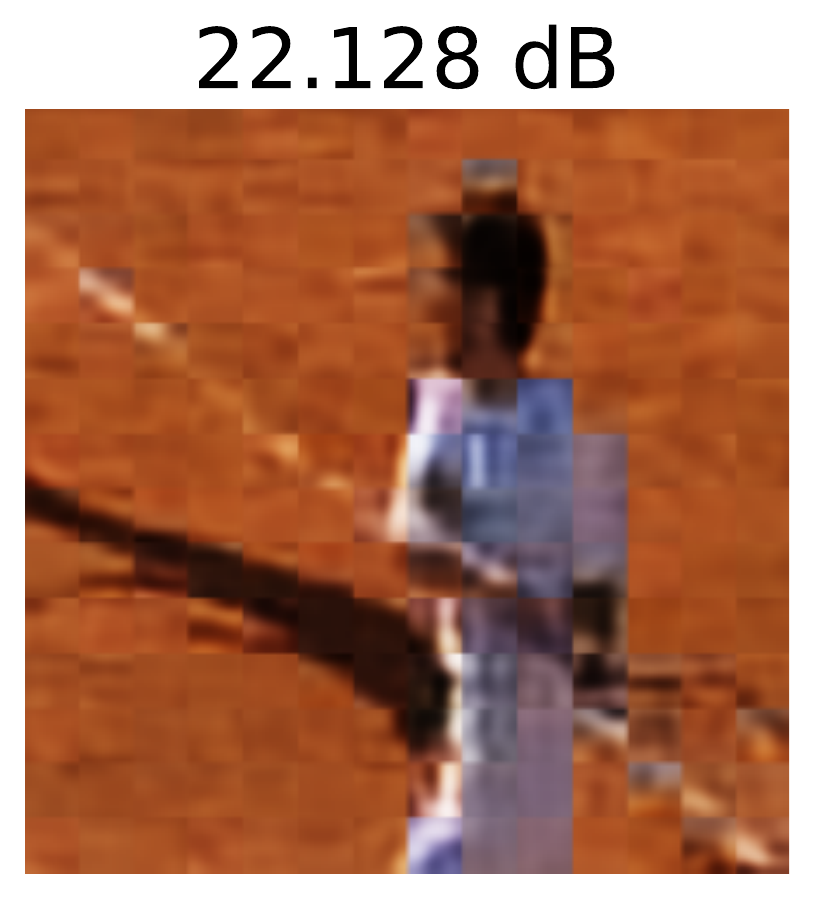}
        \caption{$\gamma = 10^{-6}$}
        \end{subfigure}
        \begin{subfigure}{.18\textwidth}
        \includegraphics[width=0.99\linewidth]{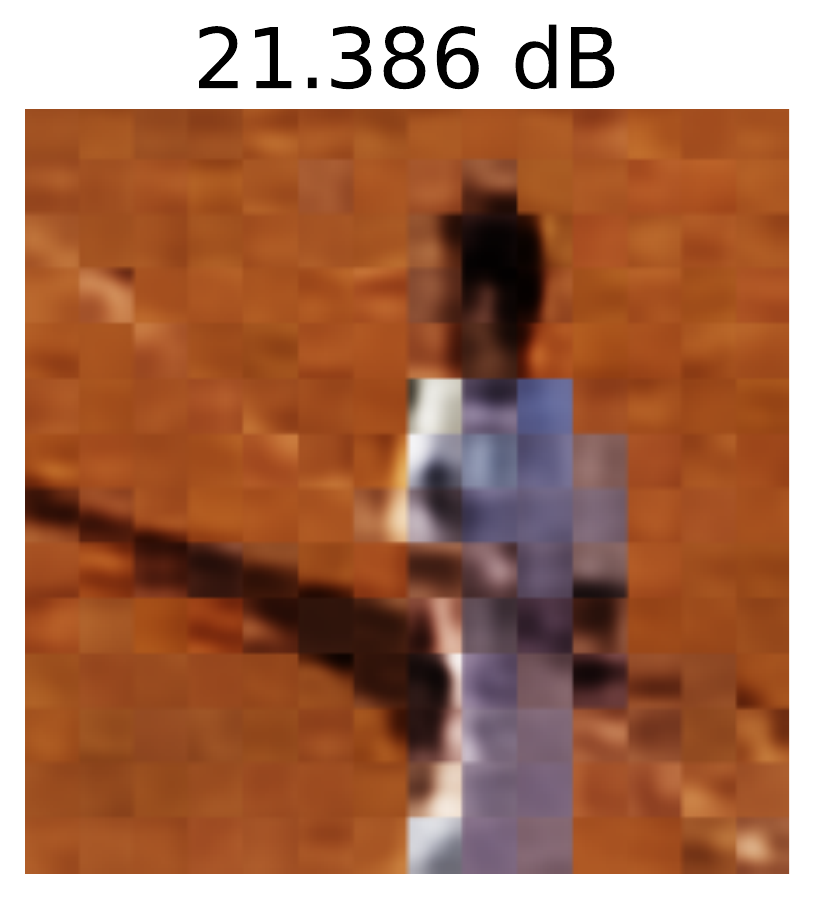}
        \caption{$\gamma = 5 \times 10^{-7}$}
        \end{subfigure}
        \caption{\small Input image and sample reconstructions with PSNR (in dB) values from the validation set for transformer-based decoder with different reconstruction loss weights $\gamma$.}
        \label{fig:recon_sample_loss_sweep}
    \end{figure}

    For both decoder types, marginal gains are only possible if the reconstruction loss weight $\gamma$ is chosen appropriately such that the reconstructions are imperfect. Optimizing for perfect accuracy by using larger weights significantly degrades the multi-label classification evaluation performance. 
    The multi-label classification mAP values for various $\gamma$ in Fig. \ref{fig:recon_loss_sweep} highlight this fact. Reconstruction samples from the validation set in Fig. \ref{fig:recon_sample_loss_sweep} show how accuracy changes with respect to the loss weight.
    
\end{document}